\documentclass[runningheads,a4paper]{llncs}

\usepackage{amssymb}
\usepackage{floatflt}
\usepackage{times}
\usepackage{multirow}
\usepackage{hyperref}
\usepackage[linesnumbered]{algorithm2e}
\usepackage[pdftex]{graphicx}
\usepackage[latin1]{inputenc}
\usepackage{subcaption}

\usepackage{url}
\newcommand{\keywords}[1]{\par\addvspace\baselineskip
\noindent\keywordname\enspace\ignorespaces#1}

\begin{document}

\mainmatter  

\title{ExpertBayes: Automatically refining manually built Bayesian networks}

\titlerunning{ExpertBayes}

%
%
\author{Ezilda Almeida\inst{1} \and
Pedro Ferreira\inst{2} \and Tiago Vinhoza\inst{3}\and In\^es Dutra\inst{2}\\
Jingwei Li\inst{4} \and Yirong Wu\inst{4} \and Elizabeth Burnside\inst{4}
}
\authorrunning{Almeida et al.}

\institute{CRACS INESC TEC LA, 
\and
Department of Computer Science, University of Porto
\and
Instituto de Telecommunica\c c\~oes, IT-Porto
\and
University of Wisconsin, Madison, USA
}

\maketitle

\begin{abstract}
Bayesian network structures are usually built using only the data and
starting from an empty network or from a na\"ive Bayes structure. Very
often, in some domains, like medicine, a prior structure knowledge is
already known. This structure can be automatically or manually refined
in search for better performance models. In this work, we take
Bayesian networks built by specialists and show that minor
perturbations to this original network can yield better classifiers
with a very small computational cost, while maintaining most of the
intended meaning of the original model.
\end{abstract}

\section{Introduction}

Bayesian networks are directed acyclic graphs that represent
dependencies between variables in probabilistic models.  In these
networks, each node represents a variable of interest and the edges
may represent causal dependencies between these variables. A Bayesian
network encodes the
Markov assumption that each variable is independent of its
non-descendants, given just its parents. Each node (variable) is
associated with a conditional probability table. 

When used for knowledge representation, a network is simply a
graphical model that represents relations among variables. This
graphical model can be learned from data or can be manually built. In
the latter case, the network encodes the knowledge of an expert
and can serve as a basis for the construction of new networks. When
learned only from data, the final graphical model (network structure) may
not have a meaning for a specialist in the domain defined by the data. 

In this work, we aim to gather the advantages of manual construction
with the advantages of automatic construction, using ExpertBayes, a
system that implements an algorithm that can refine previously built
networks. ExpertBayes allows for (1) reducing the computational costs
involved in building a network only from the data, (2) embedding
knowledge of an expert in the newly built network and (3) manual
building of fresh new graphical representations. The main ExpertBayes
algorithm is random and implements 3 operators: insertion, removal and
reversal of edges. In all cases, nodes are also chosen randomly.

Our expert domains are prostate cancer and breast cancer. We used
graphical models manually built by specialists as starting
networks. Parameters are learned from the data, but can also be given
by the specialists. We compare the performance of our original
networks with the best network found using our random
algorithm. Results are validated using 5-fold cross-validation. For
different threshold values, results, both in the training and test
sets, show that there is a statistically significant difference
between the original network and the newly built networks. As far as
we know, this is the first implementation of an algorithm capable of
constructing Bayesian networks from prior knowledge in the form of a
network structure. Previous works considered as initial network a
na\"ive Bayes or an empty
network~\cite{Hall:2009:WDM:1656274.1656278,bnlearn1,bnlearn2,Chan:2004:SAB:1036843.1036852}. As
far as we know, the R package \texttt{deal}~\cite{Bottcher03deal} is
the only one that refines previous Bayesian structures, but our
attempts to make it work were not successful, since the parameters
computed for the new networks were not interpretable. We then decided
to implement our own algorithm. 

One important aspect of ExpertBayes is
that it makes small perturbations to the original model thus
maintaining its intended meaning. Besides refining pre-defined networks,
ExpertBayes is interactive. It allows users to play with the network
structure which is an important step in the integration of expert
knowledge to the automatic learning process.

\section{ExpertBayes: refining expert-based Bayesian networks}

Most works in the literature that discuss methods for learning the
structure of Bayesian networks focus on learning from an empty network
or from data. However, in some domains, it is common to find Bayesian
models manually built by experts, using tools such as GeNIe (a
modeling environment developed by the Decision Systems Laboratory of
the University of Pittsburgh, available at
\url{http://genie.sis.pitt.edu}), Netica
(\url{https://www.norsys.com/netica.html}) or the WEKA Bayes
editor~\cite{Hall:2009:WDM:1656274.1656278}. Having an initial model
brings at least two advantages: (1) from the point of view of the
specialist, some expert knowledge has already being embedded to the
model, with meaningful correlations among variables, (2) from the
point of view of the structure learning algorithm, the search becomes
less costly, since an initial structure is already known. In fact, in
other areas, it is very common to use previous knowledge to reduce the
search space for solutions. One classical example is the comb-like
structure used as initial seed for DNA reconstruction algorithms
based on Steiner minimum trees. In the past, the protein structure was
searched for from an empty initial structure~\cite{Smith}. The
discovery that most protein structures in the nature had a comb-like
shape reduced the algorithm cost allowing to solve much bigger
problems~\cite{Mondaini:2005:ISP:1106327.1106559}.

ExpertBayes uses a simple, yet efficient algorithm to refine the
original network. This algorithm is shown in
Figure~\ref{alg:novaRede}. It reads the initial input network and
training and test sets. It then uses a standard method to initialize
the probability tables, by counting the case frequency of the training
set for each table entry. Having the prior network and conditional
probability tables, the algorithm makes small perturbations to the
original model. It first chooses a pair of nodes, then it randomly
chooses to add, remove or revert an edge.  If the operation is to add
an edge, it will randomly choose the edge direction. Operations are
applied if no cycle is produced. At each of these steps, conditional
probability tables are updated, if necessary, i.e., if any node
affected belongs to the Markov blanket of the classifier node. A score
of the new model is calculated for the training set and only the best
pair network/score is retained when the repeat cycle ends. This best
network is then applied to the test set (last step, line 20 of the
algorithm). A global score metrics is used, the number of correctly
classified instances, according to a threshold of 0.5.

\begin{algorithm}[htbp]
\caption{ExpertBayes}

\SetAlgoLined
\KwData{\\ OriginalNet, \emph{// initial network structure}\;
        \\ Train \emph{// training set}\;
        \\ Test \emph{// test set}}
\KwResult{\\ scoreTrain \emph{// scores in the training set for BestNet}
          \\ scoreTest \emph{// scores in the test set for BestNet}
          \\ BestNet \emph{// best scored network on Train}}
\BlankLine
Read OriginalNet\;
Read Train and Test sets\;
BestNet = OriginalNet\;
Learn parameters for OriginalNet from training set\;
\Repeat{N iterations using OriginalNet and Train}{
  Randomly choose a pair of nodes $N_1$ and $N_2$\;
  \eIf{there exists an edge between $N_1$ and $N_2$}
     {randomly choose: revert or remove}
     {choose add operation\; randomly choose edge direction}
  Apply operation to OriginalNet obtaining NewNet\;
  Rebuild necessary CPT entries, if necessary\;
  Compute scoreTrain of the NewNet\;
  \If{scoreTrain NewNet \textgreater\ scoreTrain BestNet}{BestNet = NewNet}
}
Apply BestNet to Test and compute scoreTest\;

\label{alg:novaRede}
\end{algorithm}

The modifications performed by ExpertBayes are always over the
original network. This was strategically chosen in order to cause a minimum
interference on the expert knowledge represented in the graphical
model. ExpertBayes has also the capability of creating a new network
only from the data if the user has no initial network to provide.

\section{Materials and Methods}

The manual construction of a Bayesian network can be tedious and
time-consuming. However, the knowledge encoded in the graphical model
and possibly in the prior probabilities is very valuable. We are lucky
enough to have two of these networks. One was built for the domain of
prostate cancer and the second one was built for breast cancer. 

In the prostate cancer domain, variables were
collected~\cite{Sarabando} taking into account three different moments
in time: (1) during a medical appointment, (2) after auxiliary exams
are performed and (3) five years after a radical prostatectomy. Such
variables are age, weight, family history, systolic and diastolic
arterial blood pressure, hemoglobin rate, hypoecogenic nodules,
prostate specific-antigen (psa), clinical status, doubling time PSA, prostate
size, among others. Five years after the surgery, we assess morbidity for those
patients.

The data for breast cancer was collected from patients of the
University of Wisconsin Medical Hospital. Mammography features were
annotated according to the BI-RADS (Breast Imaging and Data Reporting
System)~\cite{BIRADS}. These include breast density, mass density,
presence of mass or calcifications and their types, architectural
distortion, among others. One variable indicates the diagnostic and
can have values malignant or benign, to indicate the type of finding.

A third set of data was used, also with mammographic features from the
University of Wisconsin Medical Hospital, but with a different set of
patients and a smaller number of variables.

\subsection{Original Bayesian Networks} 

\begin{figure*}[hbtp]
\centerline{\includegraphics[width=8cm,height=5cm]{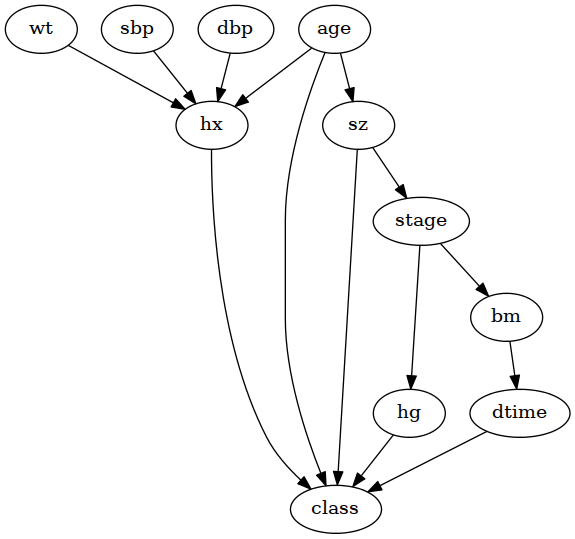}}
\caption{Original Network Model for Prostate Cancer}
\label{fig:pros}
\end{figure*}

\begin{figure*}[hbtp]
\centerline{\includegraphics[width=10cm,height=4cm]{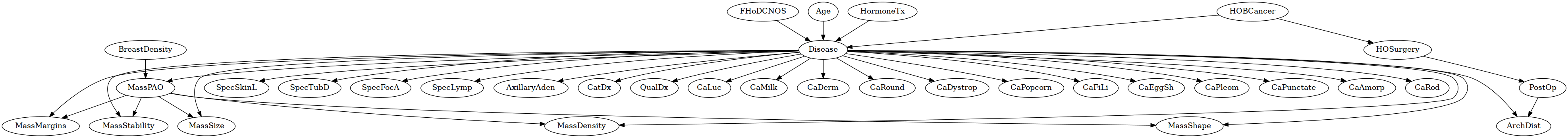}}
\caption{Original Network Model for Breast Cancer (1)}
\label{fig:bc}
\end{figure*}

Two of our networks were built by specialists while the third one was
built by us. The Bayesian networks built by our specialists are shown
in Figures~\ref{fig:pros}~\cite{Sarabando}
and~\ref{fig:bc}~\cite{BurnsideDavis09}.

\begin{figure*}[hbtp]
\centerline{\includegraphics[width=10cm,height=3cm]{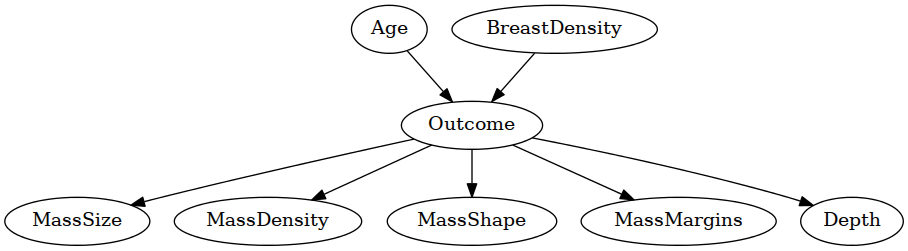}}
\caption{Original Network Model for Breast Cancer (2)}
\label{fig:bc2}
\end{figure*}

We call them Original Networks.  Both of them were built by
specialists in prostate cancer and breast cancer using high risk and
low risk factors mentioned in the literature and their own
experience. Prior probabilities are taken from the training data.  The
class variable for the breast cancer data is CatDx. In other words,
the classifying task is to predict a malignant or benign finding. The
class variable for the prostate cancer data is the life expectancy
five years after the surgery, called class in Figure~\ref{fig:pros}.

The third network was also manually built using the model of
Figure~\ref{fig:bc} as a basis, but with a smaller set of features used in
another work~\cite{IJDMB}. The class variable is Outcome with values
malignant or benign.

\subsection{Datasets}

The characteristics of the datasets used are shown in
Table~\ref{tab:chars}. The three of them have only two classes. For
Breast Cancer (1) and Breast Cancer (2), the Pos column indicates the
number of malignant cases and the Neg column indicates the number of
benign cases. For Prostate Cancer, the Pos column indicates the number
of patients that did not survive 5 years after surgery.

\begin{table}[htbp]
\begin{center}
\begin{tabular}{l||c|c|c|c} \hline
Dataset         &  Number of Instances  & Number of Variables & Pos & Neg
\\ \hline 
Prostate Cancer   &  496                  &  11                 & 352 & 144 \\
Breast Cancer (1) &  100                  &  34                 &  55 & 45 \\ 
Breast Cancer (2) &  241                  &   8                 &  88 & 153 \\ \hline
\end{tabular}
\caption{Datasets Descriptions}
\label{tab:chars}
\end{center}
\end{table}

The dataset for Prostate Cancer is available from
\url{http://lib.stat.cmu.edu/S/Harrell/data/descriptions/prostate.html}~\cite{ProstData}. 

For each one of the datasets, variables with numerical values were
discretized according to reference values in the domain (for example,
variables such as age and size are discretized in intervals with a clinical
meaning). The same discretized datasets were used with all algorithms.

\subsection{Methodology}

We used 5-fold cross-validation to train and test our models. We
compared the score of the original network with the score of
ExpertBayes. We also used WEKA~\cite{Hall:2009:WDM:1656274.1656278} to
build the network structure from the data with the K2~\cite{k2} and
TAN~\cite{TAN} algorithms. K2 is a greedy algorithm that, given an
upper bound to the number of parents for a node, tries to find a set
of parents that maximizes the likelihood of the class variable. TAN
(Tree Augmented Na\"ive Bayes) starts from a na\"ive Bayes structure
where the tree is formed by calculating the maximum weight spanning
tree using Chow and Liu algorithm~\cite{ChowLiu}. In practice, TAN
generates a tree over na\"ive Bayes structure, where each node has at
most two parents, being one of them the class variable. We ran both
algorithms with default values and both start from a na\"ive Bayes
structure. The best networks found are shown and contrasted to the
original network and to the network produced by ExpertBayes.

\section{Results}

In this Section, we present the results measured using CCI (percentage
of Correctly Classified Instances) and Precision-Recall
curves. Precision-Recall curves are less sensitive to imbalanced data
which is the case of our datasets. We
also discuss about the quality of the generated networks.

\subsection{Quantitative Analysis}

\subsubsection{CCI}

Table~\ref{tab:cci} shows the results (Correctly Classified Instances
- CCI) for each test set and each network. Results are shown in
percentages and are macro-averaged across the five folds. All results are
shown for a probability threshold of 0.5. 

\begin{table*}[htbp]
\begin{center}
\begin{tabular}{l||c|c|c|c} \hline
    Dataset         & Original & ExpertBayes & WEKA-K2 & WEKA-TAN \\ \hline
Prostate Cancer     & 74       & 76          & 74      & 71       \\ 
Breast Cancer (1)   & 49       & 63          & 59      & 57       \\
Breast Cancer (2)   & 49       & 64          & 80      & 79       \\ \hline
\end{tabular}
\caption{CCI test set - averaged across 5-folds}
\label{tab:cci}
\end{center}
\end{table*}

For the Prostate Cancer data, ExpertBayes is better than WEKA-TAN with
$p<0.01$. The difference is not statistically significant between the
ExpertBayes and the Original Network results and ExpertBayes and
WEKA-K2.

With $p<0.004$, for Breast Cancer (1), ExpertBayes produces better
results than the Original Network (63\% CCI against 49\% CCI of the
original network). With the same p-value, ExpertBayes (63\% CCI) is
also better than WEKA-K2 (59\%). With $p<0.002$, ExpertBayes is better
than WEKA-TAN (57\%).

For Breast Cancer (2), WEKA-K2 is better than ExpertBayes with
$p<0.003$. WEKA-TAN is also better than ExpertBayes with
$p<0.008$. ExpertBayes is only better than the
original network, with $p<0.009$.



Recall that these results are achieved with a threshold of 0.5.

\subsubsection{Precision-Recall Analysis}

\begin{figure*}[hbtp]
      \begin{subfigure}{0.33\linewidth}
            \includegraphics[width=8cm]{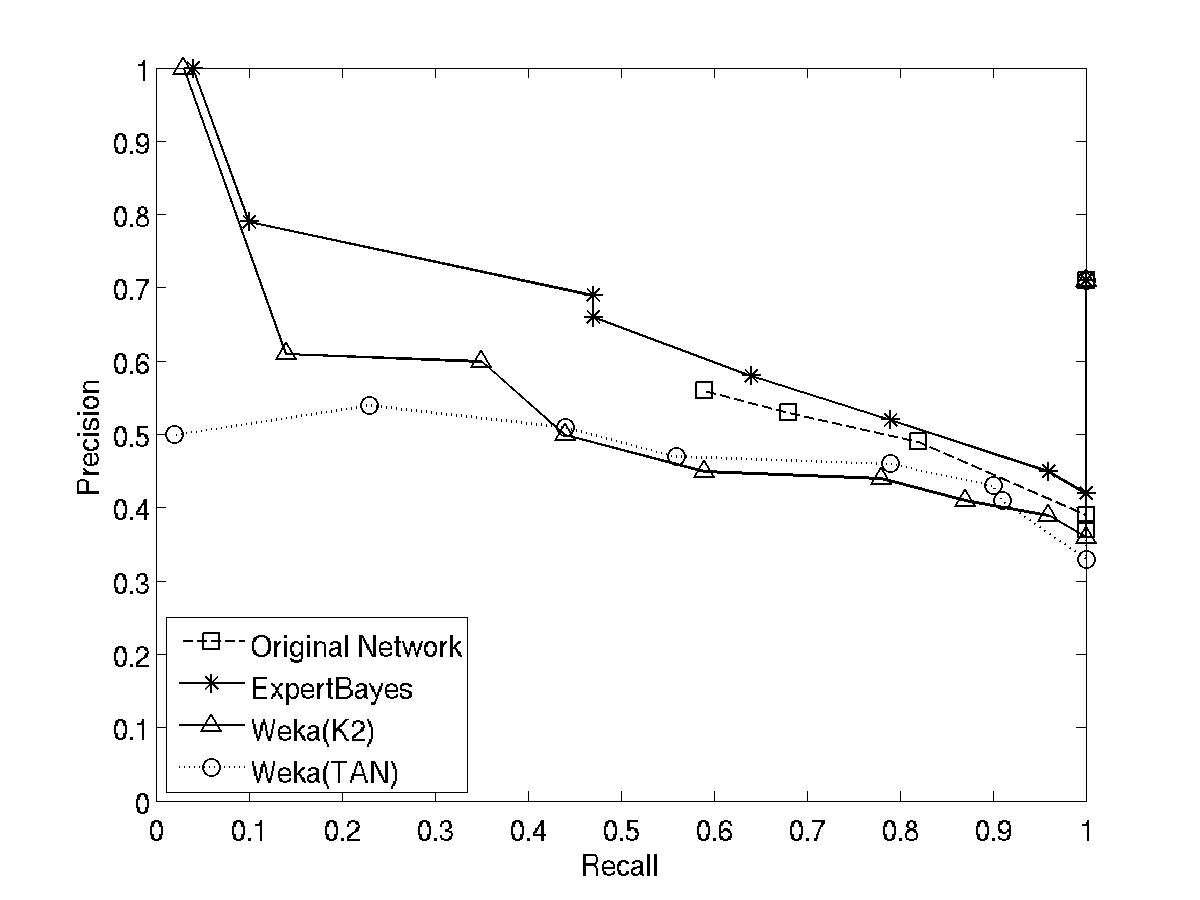}
            \caption{Prostate}
      \end{subfigure}

      \begin{subfigure}{.33\linewidth}
            \includegraphics[width=8cm]{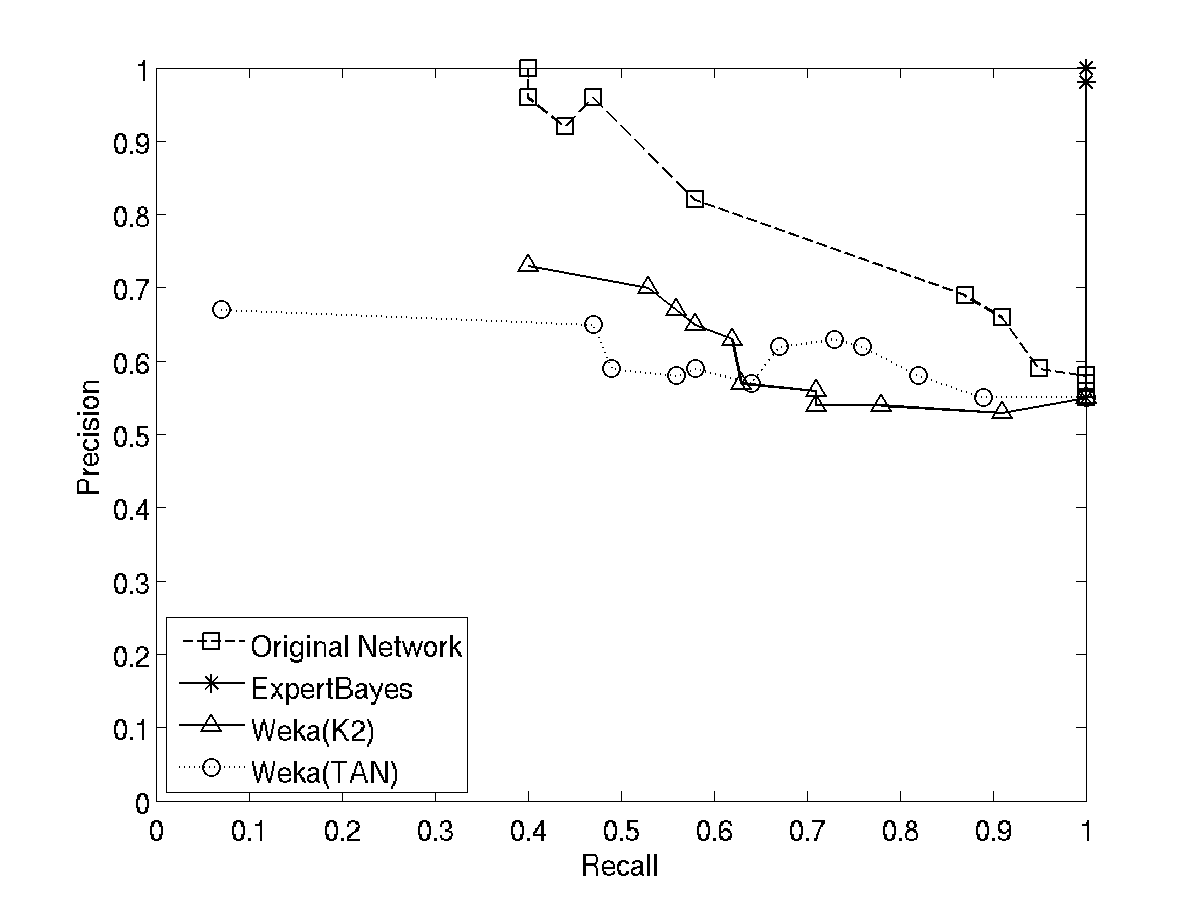}
            \caption{Breast Cancer (1)}\label{fig:bc1PR}
      \end{subfigure}

      \begin{subfigure}{0.33\linewidth}
            \includegraphics[width=8cm]{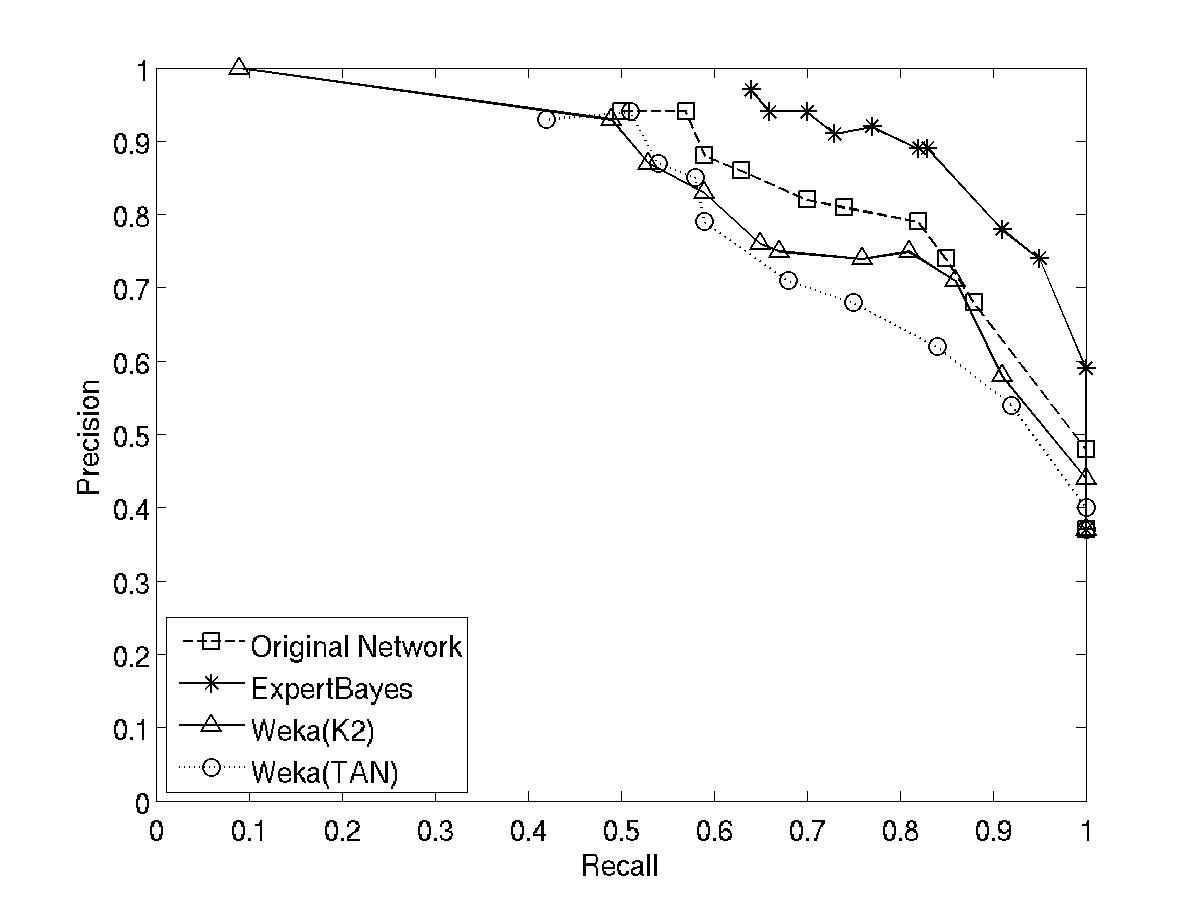}
            \caption{Breast Cancer (2)}\label{fig:bc2PR}
      \end{subfigure}
\caption{Precision-Recall Curves for various thresholds}
\label{fig:PR}
\end{figure*}

Instead of looking only at CCI with a threshold value of 0.5, we also
plotted Precision-Recall curves.
Figure~\ref{fig:PR} shows
the curves for the three datasets. Results are shown
for the test sets after cross-validation. We used values of 0.02 and 0.1
(threshold values commonly used in clinical practice for mammography
analysis) and also varied the thresholds in the interval 0.2-1.0. 

The baseline precision for the three datasets are: 71\% for Prostate
Cancer, 55\% for Breast Cancer (1) and 37\% for Breast Cancer
(2). These baseline values correspond to classifying every case as
belonging to one class. For Breast Cancer (1) and Breast Cancer (2),
this class is malignant. For Prostate Cancer, the class is not
survival.

The first important conclusion we can take from these curves is that
ExpertBayes is capable of improving Precision over the other models,
at the same Recall level. In practice, this means that a smaller
number of healthy patients will be sent to inconvenient procedures in
the case of breast cancer analysis and a smaller number of patients will
have a wrong prognostic of not survival after 5 years of surgery for
the Prostate cancer analysis.

The second conclusion we can take is that expert-based models applied
to data produce better performance than the traditional network
structures built only from the data. This means that expert knowledge
is very useful to help giving an initial efficient structure. This
happened to all datasets.

A third conclusion we can take is that a small set of features can
have a significant impact on the performance of the classifier. If we
compare Figure~\ref{fig:bc1PR} with Figure~\ref{fig:bc2PR}, all
classifiers for Breast Cancer (2) outperform the classifiers of Breast
Cancer (1). This may indicate that to prove malignancy, an expert need
to look at a fewer number of features.

One caveat, though, needs to be avoided. If we look at the performance
of the model produced by ExpertBayes for Breast Cancer (1), this is
perfect for a given threshold, with maximum Recall and maximum
Precision. This can happen when variables are highly correlated as is
the case of Disease and CatDx. In our experiments, WEKA did not
capture this correlation because the initial network used is a na\"ive
structure (no variable ever has an edge directed to the class
variable). As we allow edge reversal, the best network found is
exactly one where Disease has an edge directed to the CatDx class
variable. However, this is an excellent opportunity to the interactive
aspect of ExpertBayes, since the expert now can notice that this happens
and can remove one of the nodes or prevent the reverted correlation from
happening. 

\subsection{Bayes Networks as Knowledge Representation}

\begin{figure*}[hbtp]
      \begin{subfigure}[b]{.45\linewidth}
            \includegraphics[width=10cm]{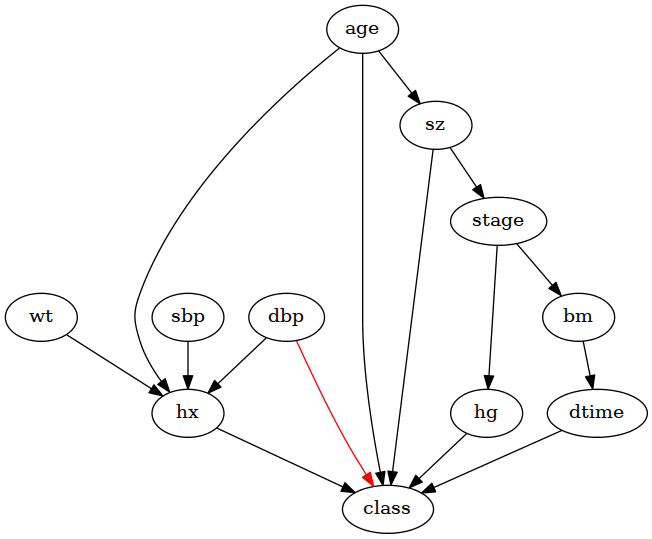}
            \caption{ExpertBayes}\label{fig:prosEB}
      \end{subfigure}

\vspace*{0.5cm}
      \begin{subfigure}[b]{0.45\linewidth}
            \includegraphics[width=10cm]{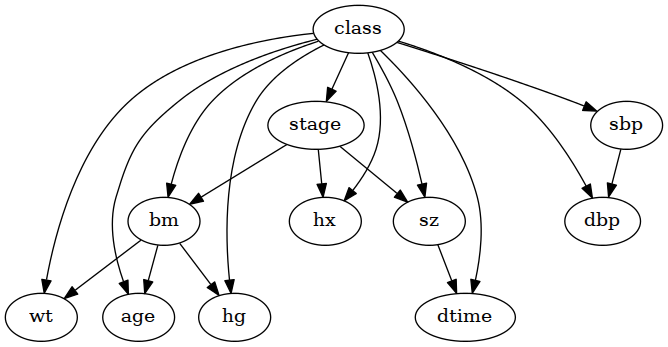}
            \caption{WEKA-TAN}
      \end{subfigure}

\vspace*{0.5cm}
      \begin{subfigure}[b]{0.45\linewidth}
            \includegraphics[width=10cm,height=3cm]{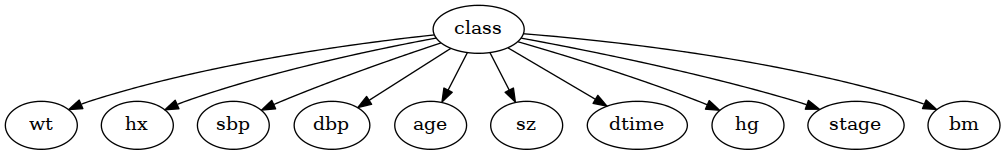}
            \caption{WEKA-K2}
      \end{subfigure}\hfill
\caption{Best Models for Prostate Cancer}
\label{fig:bestProstate}
\end{figure*}

\begin{figure*}[hbtp]
      \begin{subfigure}[b]{.45\linewidth}
            \includegraphics[width=14cm,height=4cm]{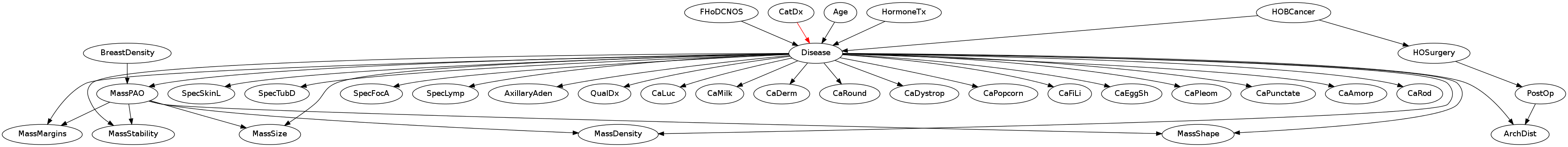}
            \caption{ExpertBayes}\label{fig:bc1EB}
      \end{subfigure}

\vspace*{0.5cm}
      \begin{subfigure}[b]{0.45\linewidth}
            \includegraphics[width=14cm,height=5cm]{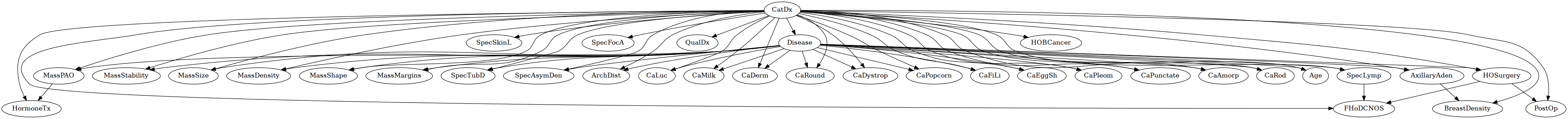}
            \caption{WEKA-TAN}
      \end{subfigure}

\vspace*{0.5cm}
      \begin{subfigure}[b]{0.45\linewidth}
            \includegraphics[width=14cm,height=2.5cm]{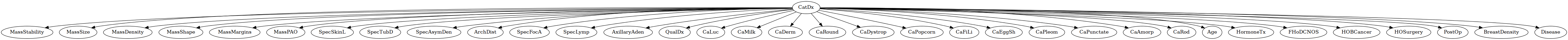}
            \caption{WEKA-K2}
      \end{subfigure}\hfill
\caption{Best Models for Breast Cancer (1)}
\label{fig:bestMammo}
\end{figure*}

\begin{figure*}[hbtp]
      \begin{subfigure}[b]{.45\linewidth}
            \includegraphics[width=10cm]{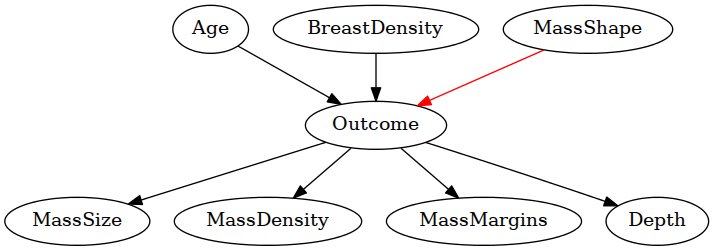}
            \caption{ExpertBayes}\label{fig:bc2EB}
      \end{subfigure}

\vspace*{0.5cm}
      \begin{subfigure}[b]{0.45\linewidth}
            \includegraphics[width=10cm]{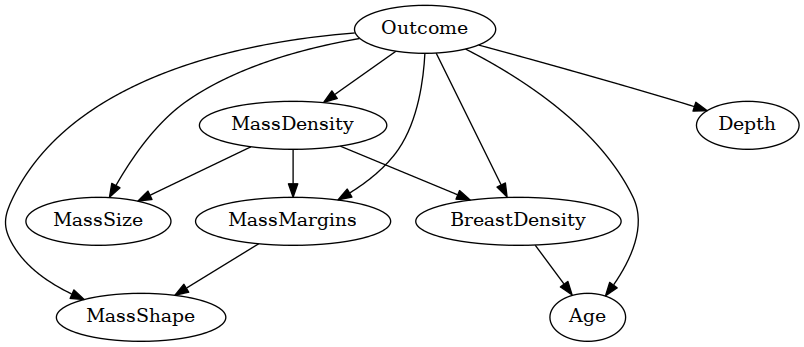}
            \caption{WEKA-TAN}
      \end{subfigure}

\vspace*{0.5cm}
      \begin{subfigure}[b]{0.45\linewidth}
            \includegraphics[width=10cm,height=3cm]{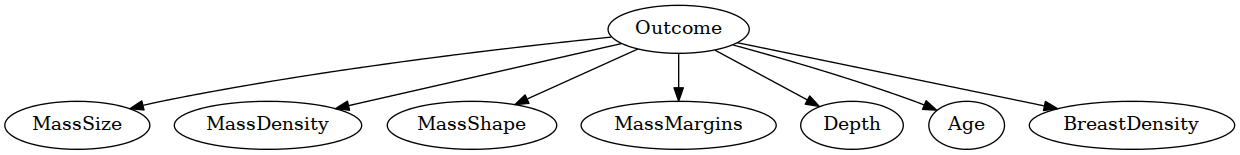}
            \caption{WEKA-K2}
      \end{subfigure}\hfill
\caption{Best Models for Breast Cancer (2)}
\label{fig:bestMammo2}
\end{figure*}

Examples of the best networks produced by ExpertBayes and WEKA-K2 and
WEKA-TAN are shown in Figure~\ref{fig:bestProstate} for Prostate
Cancer and in Figures~\ref{fig:bestMammo}
and~\ref{fig:bestMammo2} for Breast Cancer (1) and Breast Cancer (2).

The best networks produced by ExpertBayes maintain the original
structure with its intended meaning and show one single modification
to the original model by adding, removing or reversing an edge. For
example, for Prostate Cancer, Figure~\ref{fig:prosEB}, a better
network was produced that shows a relation between the diastolic blood
pressure (dbp) and the class variable. It remains to the specialist to
evaluate if this has some clinical meaning. For Breast Cancer (1), the
best network is found when a correlation is established between
MassMargins and the class variable (Figure ~\ref{fig:bc1EB}). It is well known
from the literature in breast cancer that some BI-RADS factors are
very indicative of malignancy and MassMargins is one of them. For
Breast Cancer (2) (Figure~\ref{fig:bc2EB}, the best network produced
by ExpertBayes has an added edge between MassShape and Outcome,
indicating that besides Age and BreastDensity, MassShape has also some
influence on the class variable.

Results produced with the WEKA tool show networks very different from
the ones built by experts. This was expected since the model is built
only from the data and not all possible networks are searched for due
to the complexity of searching for all possible models. The K2
algorithm found that the best model for all datasets was the na\"ive
Bayes model. Both models produced using K2 and TAN convey another
meaning to the specialist that is quite different from the initial
intended meaning.  This happened with all networks produced by WEKA,
for both datasets.

\section{Conclusions}

We implemented a tool that can allow the probabilistic study of
manually built bayesian networks. ExpertBayes is capable of taking as
input a network structure, learn the initial parameters, and iterate,
producing minor modifications to the original network structure,
searching for a better model while not interfering too much with the
expert knowledge represented in the graphical model. ExpertBayes makes
small modifications to the original model and obtain better results
than the original model and better than models learned only from the
data. Building a Bayesian network structure from the data or from a
na\"ive Bayes structure is very time-consuming given that the search
space is combinatorial. ExpertBayes takes the advantage of starting
from a pre-defined structure. In other words, it does not build the
structure from scratch and takes advantage of expert knowledge to
start searching for better models. Moreover, it maintains the basic
structure of the original network keeping its intended
meaning. ExpertBayes is also an interactive tool with a graphical user
interface (GUI) that allows users to play with their models thus
exploring new structures that give rise to a search for other
models. We did not stress this issue in this work as our focus was on
showing that ExpertBayes can refine well pre-defined models. Our main
goal for the future is to improve the algorithm in order to have
better prediction performance, possibly using more and quality data
and different search and parameter learning methods. We also intend to
embed in ExpertBayes a detection of highly correlations that exist
among variables to warn the expert. If this
is done before learning we could avoid producing unnecessary
interactions between the user and the system.

\section*{Acknowledgements}

This work was partially funded by projects ABLe (PTDC/ EEI-SII/ 2094/
2012), ADE (PTDC/EIA-EIA/121686/2010) and QREN NLPC Coaching
International 38667 through FCT, COMPETE and FEDER and R01LM010921:
Integrating Machine Learning and Physician Expertise for Breast Cancer
Diagnosis from the National Library of Medicine (NLM) at the National
Institutes of Health (NIH).

\makeatletter
\renewcommand{\@biblabel}[1]{\hfill #1.}
\makeatother

\bibliographystyle{splncs03}
\bibliography{ecml2014}

\end{document}